\documentclass[conference]{IEEEtran}
\IEEEoverridecommandlockouts

\usepackage{cite}
\usepackage{enumitem}
\usepackage{amsmath,amssymb,amsfonts}
\usepackage{algorithm}
\usepackage{algorithmic}
\usepackage{graphicx}
\usepackage{textcomp}
\usepackage{xcolor}
\usepackage{forest}
\usetikzlibrary{trees,positioning,shapes,shadows,arrows.meta, calc}
\usepackage{tikz}
\usepackage{adjustbox} 
\usepackage{url}

\makeatletter
\newcommand{\linebreakand}{%
  \end{@IEEEauthorhalign}
  \hfill\mbox{}\par
  \mbox{}\hfill\begin{@IEEEauthorhalign}
}
\makeatother

\begin{document}

\title{From Noise to Nuance: Advances in Deep Generative Image Models}

\author{
    \IEEEauthorblockN{
        Benji Peng\textsuperscript{*, a, b},
        Chia Xin Liang \textsuperscript{c},
        Ziqian Bi\textsuperscript{d}, 
        Ming Liu\textsuperscript{e},
        Yichao Zhang\textsuperscript{f}, \\
        Tianyang Wang\textsuperscript{g},
        Keyu Chen\textsuperscript{a},
        Xinyuan Song \textsuperscript{h}
        Pohsun Feng\textsuperscript{i},
    }
    \IEEEauthorblockA{
        \textsuperscript{a}Georgia Institute of Technology, USA
    }
    \IEEEauthorblockA{
        \textsuperscript{b}AppCubic, USA
    }
    \IEEEauthorblockA{
        \textsuperscript{c}JTB Technology Corp., ROC
    }
    \IEEEauthorblockA{
        \textsuperscript{d}Indiana University, USA
    }
    \IEEEauthorblockA{
        \textsuperscript{e}Purdue University, USA
    }
    \IEEEauthorblockA{
        \textsuperscript{f}The University of Texas at Dallas, USA
    }
    \IEEEauthorblockA{
        \textsuperscript{g}University of Liverpool, UK
    }
    \IEEEauthorblockA{
        \textsuperscript{h}Emory University, USA
    }
    \IEEEauthorblockA{
        \textsuperscript{i}National Taiwan Normal University, ROC
    }
    \IEEEauthorblockA{
        *Corresponding Email: benji@appcubic.com
    }
}

\maketitle


\begin{abstract}
Deep learning-based image generation has undergone a paradigm shift since 2021, marked by fundamental architectural breakthroughs and computational innovations. Through reviewing architectural innovations and empirical results, this paper analyzes the transition from traditional generative methods to advanced architectures, with focus on compute-efficient diffusion models and vision transformer architectures. We examine how recent developments in Stable Diffusion, DALL-E, and consistency models have redefined the capabilities and performance boundaries of image synthesis, while addressing persistent challenges in efficiency and quality. Our analysis focuses on the evolution of latent space representations, cross-attention mechanisms, and parameter-efficient training methodologies that enable accelerated inference under resource constraints. While more efficient training methods enable faster inference, advanced control mechanisms like ControlNet and regional attention systems have simultaneously improved generation precision and content customization. We investigate how enhanced multi-modal understanding and zero-shot generation capabilities are reshaping practical applications across industries. Our analysis demonstrates that despite remarkable advances in generation quality and computational efficiency, critical challenges remain in developing resource-conscious architectures and interpretable generation systems for industrial applications. The paper concludes by mapping promising research directions, including neural architecture optimization and explainable generation frameworks.
\end{abstract}

\section{Introduction}
\subsection{Evolution of Image Generation Paradigms}

Image generation models has been through significant changes, transitioning from Generative Adversarial Networks (GANs) to diffusion models. GANs, introduced by Goodfellow et al.\cite{goodfellow2020generative}, set a new standard for generating realistic images by employing a dual-network architecture consisting of a generator and a discriminator. Despite their success, GANs faced challenges such as training instability and mode collapse, limiting their applicability in certain scenarios \cite{radford2015unsupervised}. In response to these limitations, diffusion models emerged as a robust alternative, utilizing a process that iteratively denoises data to generate high-fidelity images \cite{ho2020denoising}. These models demonstrated superior stability during training and the ability to produce diverse and detailed outputs \cite{song2020score}. Recent advancements have further enhanced diffusion models' efficiency and scalability, making them a prominent choice in image generation tasks \cite{dhariwal2021diffusion}. The shift from GANs to diffusion models marks a critical evolution in image generation paradigms, enabling more reliable and versatile applications across various domains.

Deep generative image model development has been propelled by the adoption of large-scale training methodologies. Access to extensive datasets, such as ImageNet and LAION, has enabled models to learn diverse and intricate patterns, enhancing their ability to generate high-quality images \cite{deng2009imagenet, schuhmann2021laion}. The evolution of hardware infrastructure, including powerful GPUs and TPUs, has facilitated the training of increasingly complex models \cite{jouppi2017datacenter}. Scaling laws have demonstrated that larger models tend to perform better, provided they are trained with sufficient data and computational resources \cite{kaplan2020scaling}. Innovations in distributed training techniques, such as model and data parallelism, have made it feasible to train models with billions of parameters efficiently \cite{shoeybi2019megatron, rae2021scaling}. The integration of cloud computing platforms has eased the access to the necessary computational power, allowing a broader range of researchers to engage in large-scale model training \cite{vaswani2017attention}. These factors collectively have enabled the creation of sophisticated generative models like DALL-E 2 and Stable Diffusion, which exhibit remarkable capabilities in producing nuanced and high-fidelity images \cite{ramesh2022hierarchical, saharia2022photorealistic}. The rise of large-scale training has thus been a cornerstone in the progression of deep generative image models, driving improvements in both performance and applicability.

Foundation models have revolutionized image generation by using large-scale datasets and versatile architectures to perform a wide array of tasks with minimal fine-tuning. Models such as DALL-E \cite{ramesh2021zero} and Stable Diffusion \cite{rombach2022high} demonstrates outstanding ability to generate high-quality, diverse images from textual descriptions. These models utilize transformer-based architectures and diffusion processes, enabling them to capture intricate patterns and semantics from extensive training data \cite{dhariwal2021diffusion}. The scalability of foundation models allows for improved performance across various domains, including art creation, data augmentation, and interactive design tools \cite{nichol2021improved}. Combining multimodal learning in foundation models facilitates the seamless combination of text and image inputs, enhancing the flexibility and applicability of image generation systems \cite{radford2021learning}. Recent advancements have focused on optimizing these models for efficiency and accessibility, ensuring that high-quality image generation is attainable even with limited computational resources \cite{rombach2022high}. These foundation models marks a significant milestone in image generation, providing robust and adaptable solutions that cater to a diverse range of applications.

\subsection{Current State and Challenges}

Deep generative image models have made significant strides recently, demonstrating their potential in applications like art creation, data augmentation, and simulation. Yet, several challenges still limit their broader use. Key issues include high computational demands, balancing image quality with generation speed, and complex ethical concerns. Addressing these challenges is essential for responsibly advancing generative models in practical settings. The following sections explore these obstacles, pointing out where more research and innovation are needed.

\subsubsection{Computational Scalability}

As generative image models increase in size and complexity, the demand for computational resources surges, leading to significant scalability challenges. Large models need substantial memory and processing power, making efficient training and deployment difficult \cite{karras2021alias, ho2020denoising}. Recent research has aimed to improve scalability without sacrificing performance. Techniques like model pruning, quantization, and the design of more efficient neural network architectures show promise in reducing computational costs \cite{li2024surveying}. Additionally, distributed computing and specialized hardware accelerators, such as tensor processing units (TPUs) and graphics processing units (GPUs), help manage the heavy computational load~\cite{chen2020survey}. However, balancing scalability with high-quality image generation remains an ongoing challenge, requiring further innovation and exploration.

\subsubsection{Quality-Speed Trade-offs}

Balancing image quality with inference speed remains a critical challenge in deploying deep generative models. High-fidelity image generation often demands extensive computational resources and longer processing times, making real-time applications difficult \cite{du2024age, sun2024LLMs}. Researchers have developed strategies to speed up image generation without significant quality loss. Techniques like knowledge distillation, where smaller models mimic the performance of larger ones, and lightweight architectures have proven effective in cutting inference times~\cite{li2020lightweight}. Further, advancements in algorithmic efficiency—such as optimized sampling methods and fewer iterative steps in diffusion models—help accelerate the process~\cite{lu2022dpm}. These approaches aim to balance high image quality with the speed needed for practical, scalable deployment.

\subsubsection{Ethics and Limitations}

The deployment of deep generative image models raises critical ethical challenges that demand careful management. A major concern is their potential to produce misleading or harmful content, like deepfakes, which can deceive or manipulate audiences \cite{westerlund2019emergence, laurier2024cat}. Moreover, biases in training data risk embedding and amplifying societal prejudices within generated images \cite{bianchi2023easily}. Intellectual property issues also come into play, as using copyrighted materials without permission in training datasets leads to both legal and ethical conflicts \cite{oppenlaender2022creativity}. To address these risks, it is essential to adopt strict data curation, fairness-focused training protocols, and regulatory frameworks that oversee the use and distribution of generative models \cite{peng2024jailbreaking}. Transparency in model development and deployment can further build trust, ensuring generative technologies are applied responsibly and ethically \cite{radford2021learning}. Tackling these ethical challenges is crucial for the responsible progression and societal acceptance of generative image technologies.

\section{Architectural Innovations}
Recent advances in text-to-image generative models have introduced architectural innovations to improve model capability, efficiency, and alignment with text inputs. New frameworks like transformer-based models, hybrid architectures, and refined diffusion techniques set new standards for high-resolution, contextually accurate image generation. This section examines the main architectural strategies driving these advancements, highlighting transformers, hybrid methods that blend different generative techniques, and major improvements in diffusion.

\subsection{Transformer-based Architectures}
\subsubsection{Diffusion-based Transformer Models}
\paragraph{DiT (Diffusion Transformers)}

Diffusion Transformers (DiTs) mark a major advancement in generative modeling, especially for text-to-image synthesis. Traditional diffusion models, like Denoising Diffusion Probabilistic Models (DDPMs), have mostly relied on U-Net architectures for the denoising process \cite{ho2020denoising}. DiTs, however, bring in transformer architectures and utilize their scalability and flexibility to capture complex data relationships, enabling high-quality and high-resolution images from textual descriptions \cite{peebles2023scalable, zhu2024sd}.

DiTs use a transformer to model the diffusion process and work on latent patches of input images. DiTs use a variational autoencoder (VAE) to encode images into a latent space, creating a compact, manageable representation. The transformer then processes these latent patches, capturing intricate patterns and relationships within the data. During denoising, it gradually refines these patches in a sequence of transformations, reconstructing the image with high fidelity. In the forward diffusion process, Gaussian noise is added to these latent representations across a sequence of time steps. The reverse process, guided by the transformer, iteratively denoises and reconstructs the original image by minimizing the difference between predicted noise and the actual noise added during forward diffusion \cite{peebles2023scalable}.

\paragraph{Parti Model Innovations}
The Pathways Autoregressive Text-to-Image (Parti) model, developed by Google Research, brings several advancements to text-to-image generation with a fully autoregressive approach, framing image generation as a sequential prediction task similar to language models. Parti uses transformers to capture detailed image features without relying on the iterative denoising used by diffusion models, producing images from low-res to intricate, high-res outputs. This flexibility arises from a two-stage training process: first, learning core structures at lower resolutions, then refining details at higher ones. It breaks images into patches,  efficiently handles high-res generation with minimal computational strain \cite{yu2022scaling}.

\subsubsection{Token-based Transformer Models}
\paragraph{Muse}

The Muse model by Google applies masked generative transformers to the text-to-image generation task. Unlike other models that work with continuous pixel values, Muse operates on discrete image tokens (for example, an image $x$ can be encoded into a grid of tokens $z \in \mathcal{V}^{h \times w}$, where $\mathcal{V}$ represents the vocabulary of learned codes, and $h$, $w$ are the spatial dimensions of the tokenized image), enhancing both quality and efficiency in image generation. It uses a transformer architecture, masked image modeling (MIM), trained on a masked modeling objective, where it predicts masked image tokens based on the text and the unmasked context, similar to masked language modeling in NLP. Given a text prompt $t$ and a partially masked image representation $\hat{z}$, the model learns the distribution
\begin{equation}
    p(z|\hat{z}, t) = \prod_{i \in \mathcal{M}} p(z_i|\hat{z}, t)
\end{equation}
where $\mathcal{M}$ is the set of masked positions. Muse is conditioned on text embeddings from a pre-trained language model and can interpret nuanced language prompts and generate coherent visual outputs \cite{chang2023muse}.

\begin{table*}[htbp]
\centering
\caption{CogView Model Series}
\label{table:cogview_progression}
\begin{tabular}{cp{3cm}p{12cm}}
\hline
\textbf{Model Version} & \textbf{Key Structural Changes} & \textbf{Detailed Improvements} \\ \hline
\textbf{CogView} \cite{ding2021cogview} & Base Transformer with VQ-VAE Tokenizer & 4-billion-parameter transformer using VQ-VAE to encode images as tokens. Autoregressive token prediction conditioned on text prompts, achieving state-of-the-art results on MS COCO in image coherence and detail. \\ 
\textbf{CogView2} \cite{ding2022cogview2} & Hierarchical Transformers and Parallel Autoregression & Introduces hierarchical token generation and local parallel autoregressive methods, enhancing resolution support and increasing generation speed by 10x. Enables efficient text-guided image editing and more detailed, high-resolution outputs. \\ 
\textbf{CogView3} \cite{zheng2024cogview3} & Relay Diffusion and Cascaded Super-Resolution & Adopts relay diffusion with cascaded low-to-high resolution stages, cutting down on training and inference costs. Outperforms SDXL in human evaluations with reduced inference time, offering a distilled variant achieving similar quality at 1/10th of SDXL's latency. \\ \hline
\end{tabular}
\end{table*}

\paragraph{CogView2 Developments}

The CogView model series (\textbf{Table \ref{table:cogview_progression}}) uses a token-based transformer for text-to-image generation. The original CogView model introduced in 2021 by researchers at Tsinghua University encodes images as sequences of discrete tokens via a Vector Quantized Variational AutoEncoder (VQ-VAE), effectively turning image data into a finite set of symbols manageable by a transformer. With its 4-billion-parameter transformer, CogView generates images by predicting each token in sequence, guided by the input text. This initial model showed impressive results on the MS COCO dataset, producing coherent, detailed images that surpassed GAN-based models \cite{ding2021cogview}.

CogView2 focuses on faster generation of high-resolution images. It introduced hierarchical transformers for a multi-stage generation process that produced detailed images more efficiently. CogView2 sped up the sampling process tenfold while maintaining image quality incorporating local parallel autoregressive generation. It also added text-guided image editing \cite{ding2022cogview2}. CogView3 further refined efficiency by shifting from the autoregressive approach to a relay diffusion model. It creates low-resolution images and enhances them through super-resolution stages. Relay diffusion slashed training and inference costs while maintained quality with faster generation times \cite{zheng2024cogview3}.

\subsection{Diffusion Model Breakthroughs}

\subsubsection{Denoising Diffusion Probabilistic Models}

Denoising Diffusion Probabilistic Models (DDPM) define a Markov chain of diffusion steps that gradually convert a data distribution into pure noise, then learn to reverse this process \cite{ho2020denoising}.

\paragraph{Forward Diffusion Process}
The forward process is defined as a Markov chain ($q$ represents the probability distribution in the process) that gradually adds Gaussian noise to data over $T$ timesteps:

\begin{equation}
    q(x_t|x_{t-1}) = \mathcal{N}(x_t; \sqrt{1-\beta_t}x_{t-1}, \beta_tI)
\end{equation}

where $\beta_t$ represents the noise schedule, $x_t$ is the noisy image at timestep $t$, and $x_0$ is the original image. This process can also be expressed in closed form for \textit{any} timestep $t$:

\begin{equation}
    q(x_t|x_0) = \mathcal{N}(x_t; \sqrt{\bar{\alpha}_t}x_0, (1-\bar{\alpha}_t)I)
\end{equation}

with $\alpha_t = 1 - \beta_t$ and $\bar{\alpha}_t = \prod_{i=1}^t \alpha_i$.

\paragraph{Reparameterization}
Using the reparameterization trick, a technique used in VAEs to allow the backpropagation of gradients through stochastic variables, making it feasible to train models with stochastic layers using gradient-based methods \cite{kingma2013auto}, $x_t$ can be sampled using:

\begin{equation}
    x_t = \sqrt{\bar{\alpha}_t}x_0 + \sqrt{1-\bar{\alpha}_t}\epsilon, \quad \epsilon \sim \mathcal{N}(0, I)
\end{equation}

\paragraph{Reverse Process}
The reverse process learns to denoise by estimating the noise:

\begin{equation}
    p_\theta(x_{t-1}|x_t) = \mathcal{N}(x_{t-1}; \mu_\theta(x_t, t), \Sigma_\theta(x_t, t))
\end{equation}

where the mean is parameterized as:

\begin{equation}
    \mu_\theta(x_t, t) = \frac{1}{\sqrt{\alpha_t}}\left(x_t - \frac{1-\alpha_t}{\sqrt{1-\bar{\alpha}_t}}\epsilon_\theta(x_t, t)\right)
\end{equation}

\paragraph{Training Objective}
The model is trained using a simplified variational bound:

\begin{equation}
    L_\text{simple} = \mathbb{E}_{t,x_0,\epsilon}\left[\|\epsilon - \epsilon_\theta(x_t, t)\|_2^2\right]
\end{equation}

where $t$ is uniformly sampled from $\{1,...,T\}$, $\epsilon$ is random Gaussian noise, and $\epsilon_\theta$ is the neural network predicting the noise.

\paragraph{Sampling Algorithm}
During inference, sampling is performed from the reverse process following:

\begin{algorithm}[H]
\begin{enumerate}
    \item Sample $x_T \sim \mathcal{N}(0, I)$
    \item For $t = T,...,1$:
    \begin{equation}
        x_{t-1} = \frac{1}{\sqrt{\alpha_t}}\left(x_t - \frac{1-\alpha_t}{\sqrt{1-\bar{\alpha}_t}}\epsilon_\theta(x_t, t)\right) + \sigma_t z
    \end{equation}
    where $z \sim \mathcal{N}(0, I)$ and $\sigma_t$ is the sampling noise scale
\end{enumerate}
\end{algorithm}

The variance $\Sigma_\theta(x_t, t)$ can be fixed to:
\begin{equation}
    \Sigma_\theta(x_t, t) = \beta_t I
\end{equation}

The DDPM framework enables the model to learn the reverse process of gradually denoising random Gaussian noise into meaningful data samples, establishing the foundation for modern diffusion-based generative models \cite{ho2020denoising}.

\subsubsection{Latent Diffusion Models}
Latent Diffusion Models (LDMs) marks a significant advancement in generative modeling, which addresses the inefficiencies in previous diffusion models. LDM performs the diffusion process in a compressed latent space rather than pixel space, leading to substantial improvements in both speed and memory usage while maintaining generation quality \cite{rombach2022high}.

\paragraph{Perceptual Compression Stage}
The first stage involves training an autoencoder to learn a perceptually equivalent, but computationally more efficient representation of the input data. Given an input image $x$, the encoder $\mathcal{E}$ maps it to a lower-dimensional latent space:

\begin{equation}
    z = \mathcal{E}(x) \in \mathbb{R}^{h \times w \times c}
\end{equation}

where the spatial dimensions $h$ and $w$ are typically reduced by a factor $f$ (usually 8× to 32×), and the channel dimension $c$ is optimized for information density. The autoencoder is trained using a combination of reconstruction loss and KL-regularization:

\begin{equation}
    \mathcal{L}_{\text{AE}} = \|\mathcal{D}(\mathcal{E}(x)) - x\|^2_2 + \mathcal{L}_{\text{KL}}
\end{equation}

This ensures that the latent space maintains a balance between compression and perceptual fidelity.

\paragraph{Latent Diffusion Process}
The second stage adapts the diffusion process to operate in the learned latent space. The forward process defines a Markov chain that gradually adds noise to the latent representation:

\begin{equation}
    q(z_t|z_{t-1}) = \mathcal{N}(z_t; \sqrt{1-\beta_t}z_{t-1}, \beta_tI)
\end{equation}

where $\beta_t$ represents the noise schedule. The reverse process, parameterized by a neural network $\theta$, learns to denoise by estimating:

\begin{equation}
    p_\theta(z_{t-1}|z_t, c) = \mathcal{N}(\mu_\theta(z_t, t, c), \Sigma_\theta(z_t, t, c))
\end{equation}

Here, $c$ represents arbitrary conditioning information, enabling flexible control over the generation process.

\paragraph{Cross-Attention Conditioning}
Cross-attention serves as a crucial architectural component in Latent Diffusion Models, enabling the model to incorporate conditional information (such as text prompts or image features) during the denoising process. The mechanism operates within the UNet backbone of the model, specifically in the middle layers where the diffusion process occurs.

The cross-attention layer implements the following mathematical formulation:

\begin{equation}
    \text{CrossAttention}(z, c) = \text{softmax}\left(\frac{(W_qz)(W_kc)^T}{\sqrt{d}}\right)(W_vc)
\end{equation}

where:
\begin{itemize}
    \item $z$ represents the latent features from the diffusion process
    \item $c$ represents the conditioning information
    \item $W_q, W_k, W_v$ are learnable projection matrices
    \item $d$ is the dimension of the key vectors
\end{itemize}

During the denoising process at each timestep $t$, the UNet processes the noisy latent representation $z_t$ through multiple resnet blocks. At specific layers, cross-attention is applied as follows:

\begin{equation}
    z'_t = \text{LayerNorm}(z_t + \text{CrossAttention}(z_t, c))
\end{equation}

This integration allows the model to condition the denoising process on external information. For instance, when generating images from text, the conditioning vector $c$ would be derived from the text encoder:

\begin{equation}
    c = \text{TextEncoder}(\text{text})
\end{equation}

The cross-attention mechanism effectively creates dynamic, content-dependent connections between the latent representations and the conditioning information. This allows the model to:

\begin{equation}
    p_\theta(z_{t-1}|z_t, c) = \mathcal{N}(\mu_\theta(z_t, t, c), \Sigma_\theta(z_t, t, c))
\end{equation}

where the mean and variance of the reverse process now explicitly depend on both the current noisy latent $z_t$ and the conditioning information $c$.

The attention weights computed through the softmax operation create a soft alignment between elements of the latent representation and the conditioning information. This alignment guides the denoising process by determining which aspects of the conditioning should influence different regions of the generated image. The process can be viewed as a series of guided denoising steps:

\begin{equation}
    z_{t-1} = \frac{1}{\sqrt{\alpha_t}}\left(z_t - \frac{1-\alpha_t}{\sqrt{1-\bar{\alpha}_t}}\epsilon_\theta(z_t, t, c)\right) + \sigma_t\epsilon
\end{equation}

where the noise prediction network $\epsilon_\theta$ now has access to both temporal information $t$ and conditioning information $c$ through the cross-attention mechanism.

\paragraph{Training Objective}
The model is trained using a modified objective function that operates in the latent space:

\begin{equation}
    \mathcal{L}_{\text{LDM}} = \mathbb{E}_{\mathcal{E}(x),\epsilon,t}\left[\|\epsilon - \epsilon_\theta(z_t, t)\|^2_2\right]
\end{equation}

\paragraph{Sampling Process}
During inference, the model follows a reverse process:

\begin{enumerate}
    \item Sample $z_T \sim \mathcal{N}(0, I)$
    \item For $t = T,...,1$:
        \begin{equation}
            z_{t-1} = \frac{1}{\sqrt{\alpha_t}}\left(z_t - \frac{1-\alpha_t}{\sqrt{1-\bar{\alpha}_t}}\epsilon_\theta(z_t, t)\right) + \sigma_t\epsilon
        \end{equation}
    \item Decode final result: $x = \mathcal{D}(z_0)$
\end{enumerate}

This architecture has become the foundation for many modern image generation systems, including Stable Diffusion, demonstrating that operating in a learned latent space can maintain generation quality while significantly reducing computational requirements. The flexibility of the conditioning framework and the efficiency gains have made LDMs particularly suitable for practical applications in high-resolution image synthesis.

\subsubsection{Stable Diffusion and Variants}

\begin{figure}
    \centering
    \resizebox{0.45\textwidth}{!}{ 
    \begin{tikzpicture}[
        node distance=0.1cm and 0.3cm,
        box/.style={draw, rounded corners, thick, minimum width=1.2cm, minimum height=0.8cm, fill=#1!20},
        arrow/.style={-Stealth, thick},
        font=\large
    ]
        
        \definecolor{pastelblue}{RGB}{173, 216, 230}
        \definecolor{pastelyellow}{RGB}{255, 253, 208}
        \definecolor{pastelgreen}{RGB}{196, 244, 213}
        \definecolor{pastelpink}{RGB}{255, 182, 193}
        \definecolor{pastelorange}{RGB}{255, 224, 178}

        \node[box=pastelblue] (input) {Image $x$};
        \node[box=pastelyellow, right=of input] (encoder) {Encoder $E$};
        \node[box=pastelgreen, right=of encoder] (latent) {\parbox{1.8cm}{Latent Representation $z$}};
        \node[box=pastelpink, right=of latent] (decoder) {Decoder $D$};
        \node[box=pastelblue, right=of decoder] (output) {\parbox{2.4cm}{Reconstructed Image $\hat{x}$}};

        \node[box=pastelpink, above=0.8cm of latent] (diffusion) {Diffusion Process $q(z_t|z_{t-1})$};

        \node[box=pastelorange, below=0.8cm of latent, align=center] (condition) {Conditioning \\ (e.g., Text $y$)};
        
        \draw[arrow] (input) -- (encoder);
        \draw[arrow] (encoder) -- (latent);
        \draw[arrow] (latent) -- (decoder);
        \draw[arrow] (decoder) -- (output);
        \draw[arrow, dashed] (condition) -- (latent) node[midway, left] {Cross-Attention / Concatenation};
        \draw[arrow] (latent) -- (diffusion) node[midway, left] {Noising};
        \draw[arrow] (diffusion) -- (latent) node[midway, right] {Denoising};

    \end{tikzpicture}
    }
    \caption{Stable Diffusion Model structure for training and inference.}
    \label{fig:stable_diffusion_detailed}
\end{figure}
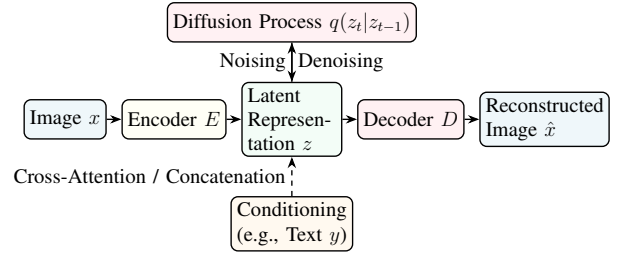

\paragraph{Stable Diffusion by Stability AI}
Stable Diffusion models, developed by Stability AI employ latent diffusion that operates in a lower-dimensional latent space rather than the high-dimensional pixel space. It significantly reduces the computational burden and enables high-resolution image generation. Stable Diffusion versions (SD1.x, SD2.x, SDXL, SD3.x) and enhanced models such as SDXL-Turbo and SD3-Turbo utilize various optimization and architectural innovations to achieve faster, high-fidelity generation with fewer inference steps. They represent a blend of computational efficiency and high-resolution output based on latent space operations, adversarial training, and architectural scaling to achieve state-of-the-art generative performance.

Stability AI has implemented several advancements to reduce sampling steps and improve fidelity. SDXL integrates a threefold increase in U-Net parameters and an improved text encoder, enhancing contextual comprehension. Additionally, multi-aspect ratio training and progressive size conditioning allow the model to generate images of varying resolutions effectively \cite{podell2023sdxl}. SDXL’s refinement model applies further denoising steps in latent space, increasing visual fidelity by reducing noise artifacts. SDXL-Turbo implements Adversarial Diffusion Distillation (ADD), which enables high-quality image generation in as few as one to four sampling steps and reduces computational demands. By integrating adversarial loss functions, SDXL-Turbo effectively mitigates common artifacts and blurriness associated with other distillation techniques, thereby improving image fidelity \cite{sauer2025adversarial}.

SD3 introduces major enhancements in architecture, efficiency, and content control, significantly advancing over its predecessors. The model transitions from a traditional U-Net to a Diffusion Transformer architecture, using separate weights for image and language representations, which improves text comprehension and prompt adherence \cite{esser2024scaling}. Enhanced text rendering now enables the generation of legible text within images, a notable improvement over prior versions. SD3 also incorporates multimodal input capabilities, allowing users to guide generation with text prompts combined with sketches or reference images, adding versatility to the creative process. Efficiency improvements include Rectified Flow sampling, which optimizes the path from noise to a clear image, as well as a novel noise schedule that samples more frequently in critical parts of the path, yielding higher-quality images. Additionally, SD3 provides scalability options, with models ranging from 800 million to 8 billion parameters, allowing users to balance computational requirements and output quality. The model also emphasizes safety, with mechanisms to prevent the generation of inappropriate content and options for artists to opt out of having their work used in training, addressing ethical concerns around content creation. Together, these upgrades establish SD3 as a powerful, flexible, and responsible tool in AI-driven image generation.

The SD3-Turbo model utilizes Latent Adversarial Diffusion Distillation (LADD), which combines adversarial training with latent space distillation to speed up inference, achieving high-quality image generation with as few as four steps. LADD applies a distilled discriminator that learns directly in the latent space, avoiding high-dimensional RGB decoding \cite{sauer2024fast}.

\begin{table*}[htbp]
\centering
\caption{Stable Diffusion Model Series by Stability AI}
\label{table:sd_progression}
\begin{tabular}{cp{3cm}p{12cm}}
\hline
\textbf{Model Version} & \textbf{Key Structural Changes} & \textbf{Detailed Improvements} \\ \hline
\textbf{SD1.x} \cite{rombach2022high} & Base Architecture & Utilizes a convolutional U-Net with attention layers for latent diffusion, enabling image synthesis with manageable parameters. CLIP ViT-L text encoder allows cross-attention, enhancing text-to-image alignment. \\ 
\textbf{SD2.x} \cite{rombach2022high} & Enhanced U-Net and Text Conditioning & Improved U-Net with enhanced parameter efficiency for high-resolution outputs. Upgraded to OpenCLIP ViT-H for text conditioning, improving prompt interpretation. Refined autoencoder optimizes perceptual loss and compute for enhanced high-frequency details. \\ 
\textbf{SDXL} \cite{podell2023sdxl} & Expanded U-Net and Dual Text Encoder & U-Net backbone scaled up by 3x, with more attention blocks and larger cross-attention context. Dual text encoders (OpenCLIP ViT-bigG and CLIP ViT-L) provide enhanced contextual understanding. Multi-aspect ratio training and a separate high-resolution refinement model support detailed backgrounds and human faces. \\ 
\textbf{SDXL Turbo} \cite{sauer2025adversarial} & Adversarial Diffusion Distillation (ADD) & Integrates ADD, combining adversarial training and score distillation, reducing sampling steps to 1-4 while maintaining fidelity. Operates in latent space, optimizing both speed and memory usage, and uses a U-Net encoder discriminator for high-resolution distillation without DINOv2 dependence. \\ 
\textbf{SD3.x} \cite{esser2024scaling} & Multimodal Diffusion Transformer (MMDiT) & Transitions from U-Net to a Diffusion Transformer, scaling parameters up to 8 billion for increased contextual comprehension. Uses separate encoders for image and language representations, enhancing text-to-image fidelity and text clarity. Uses Rectified Flow sampling for reduced generation steps and includes optimized noise scheduling for artifact reduction and improved high-resolution synthesis up to 2048x2048. \\
\textbf{SD3 Turbo} \cite{sauer2024fast} & Latent Adversarial Diffusion Distillation (LADD) & LADD achieves high-fidelity multi-aspect megapixel generation in only four steps. Distillation formulation supports diverse tasks like image editing and inpainting with improved spatial reasoning and prompt alignment. \\ \hline
\end{tabular}
\end{table*}

\paragraph{SDXL-Lightning}

SDXL-Lightning is a novel diffusion distillation method for high-quality, one-step/few-step 1024px text-to-image generation. Building upon SDXL, SDXL-Lightning combines progressive and adversarial distillation to balance generation quality and mode coverage. Progressive distillation, where a student model learns to mimic multiple steps of a teacher model, ensures the student follows the teacher's probability flow. However, using a standard mean squared error (MSE) loss in prior work \cite{salimans2022progressive} results in blurry outputs due to the student's limited capacity to capture sharp transitions in the teacher's distribution. SDXL-Lightning incorporates adversarial training at each distillation stage. A discriminator $D$, which conditions on both the current latent $x_t$ and the teacher's multi-step prediction $x_{t-ns}$ is used to guide the student $\hat{x}_{t-ns}$ to match the teacher's output distribution and maintain flow consistency. The discriminator uses the pre-trained SDXL U-Net encoder, enabling efficient operation in the latent space:
\begin{equation}
\begin{split}
D(x_t, x_{t-ns}, t, t-ns, c) = \\
\sigma( \text{head}( & d(x_{t-ns}, t-ns, c), \\  & d(x_t, t, c)))
\end{split}
\end{equation}
where $d$ represents the shared encoder and mid-block of the U-Net.  The function $\text{head}(\cdot)$ is a small neural network consisting of convolutional layers, group normalization, and SiLU activations. It takes the concatenated output of $d(x_{t-ns}, t-ns, c)$ and $d(x_t, t, c)$ as input and projects it to a single scalar value between 0 and 1 using a final sigmoid activation $\sigma(\cdot)$. This scalar represents the discriminator's confidence that the input $x_{t-ns}$ originated from the teacher network. The adversarial loss functions are:
\begin{equation}
\begin{split}
p &= D(x_t, x_{t-ns}, t, t - ns, c) \\
\hat{p} &= D(x_t, \hat{x}_{t-ns}, t, t - ns, c) \\
L_D &= -\log(p) - \log(1-\hat{p}) \\
L_G &= -\log(\hat{p})
\end{split}
\end{equation}
To further enhance semantic correctness, SDXL-Lightning relax the flow preservation constraint by finetuning with an unconditional objective.  Additionally, researchers address flaws in common diffusion schedules \cite{lin2024sdxl} by using pure noise at $t=T$ during training.  Stable training techniques, including training at multiple timesteps and adding noise to discriminator inputs, are employed, especially for one and two-step models.  Our models are available as both LoRA and full UNet weights, offering flexibility for integration and fine-tuning.

\subsubsection{Consistency Models}
A core concept in Consistency Models is the idea of a consistency function derived from the probability flow ODE of continuous-time diffusion models. Given a trajectory of a probability flow ODE, denoted as ${x_t}{t \in [\epsilon, T]}$, where $x_0$ represents the data and $x_T$ represents noise, the consistency function $f: (x_t, t) \mapsto x\epsilon$ maps any point $(x_t, t)$ on this trajectory to its origin $x_\epsilon$. A Consistency Model, denoted as $f_\theta$, aims to learn this consistency function. A key property of consistency functions, and thus Consistency Models, is self-consistency: any two points on the same probability flow trajectory should map to the same initial point, i.e., $f(x_t, t) = f(x_{t'}, t')$ for all $t, t' \in [\epsilon, T]$.

\subsubsection{Imagen Models}
Google's Imagen models are text-to-image generation systems that utilize diffusion-based architectures. The original Imagen model enables creation of high-fidelity images closely aligned with the provided text prompts by using a cascaded diffusion process, progressively generating images from low to high resolution, conditioned on textual descriptions \cite{saharia2022photorealistic}. Subsequent iterations, such as Imagen 2 \cite{google_imagen2} and Imagen 3 \cite{google_imagen3}, have introduced enhancements to the diffusion framework. These improvements include better handling of complex prompts, more accurate rendering of human features, and the ability to generate higher-resolution images. The models maintain a diffusion-based methodology, refining the process to achieve greater detail and realism in the generated images

\subsubsection{DALL-E Models}
The DALL-E model series by OpenAI emphasize the use of large-scale language-image alignment to generate coherent and contextually appropriate images from text prompts. DALL-E 1 was the first large-scale model specifically designed to generate images from text prompts. It used a transformer-based architecture, similar to models like GPT, to create images in a step-by-step manner, generating one part of the image at a time, much like forming words in a sentence \cite{ramesh2021zero}. To convert image regions into sequences of tokens, DALL-E 1 employed a discrete variational autoencoder (dVAE), allowing it to generate tokens sequentially based on the text prompt to form a complete image. For improved alignment between text and image, DALL-E 1 uses representations from CLIP, a model trained to understand text-image relationships \cite{radford2021learning} to help evaluate and rank the images. However, despite this alignment, the token-based approach limited scalability and coherence, especially when generating complex or highly detailed scenes.

DALL-E 2 generates images from text prompts using a two-stage process, starting from converting a text description into an image embedding, a sort of blueprint that captures the main visual concept. Such embedding is created with the help of CLIP, a model that aligns images and text in a shared space, making it possible to interpret complex descriptions \cite{zhang2024word}. Next, DALL-E 2 uses a diffusion model to convert this embedding into a high-quality image, gradually refining it until the final picture emerges \cite{ramesh2022hierarchical}. This setup allows DALL-E 2 to generate detailed, realistic images that closely match the given text. DALL-E 3 significantly improves the accuracy and detail of text-to-image generation, closely following complex prompts to create images that better align with user instructions. It is built upon DALL-E 2’s two-stage structure by enhancing understanding of nuanced or detailed text inputs, making it capable of handling more complex scenarios (legible text within images or rendering intricate visual elements) \cite{openai2023dalle3}. DALL-E 3 integrates with ChatGPT, which allows users to refine images interactively through conversational feedback.

\subsection{Consistency Models for Efficient Image Generation}

Consistency Models (CMs) represent a different paradigm in generative modeling, which addresses the computational inefficiencies inherent in diffusion models while maintaining high-quality generation capabilities \cite{song2023consistency}. Unlike diffusion models that require multiple denoising steps, CMs enable efficient image generation through a mathematical framework that ensures consistency across noise levels.

\subsubsection{Theoretical Foundation}
The principle of consistency ensures that the model maintains stable outputs across equivalent noise transformations. Let $\mathcal{X} \subseteq \mathbb{R}^D$ denote the data space, and consider a noise distribution $\mathcal{N}(0, \sigma_{\text{max}}^2)$. The Consistency Model is defined as a mapping $f_\theta: \mathbb{R}^D \times \mathbb{R} \to \mathbb{R}^D$ parameterized by $\theta$, which satisfies:

\begin{equation}
    f_\theta(x_t, t) = f_\theta(x_{t'}, t')
\end{equation}

where $x_t, x_{t'} \in \mathcal{X}$ represent images at different noise levels $t, t'$. This property ensures that outputs remain invariant under noise transformations, promoting structural stability during generation \cite{geng2024consistency}.

\subsubsection{Mathematical Framework}
The consistency property is enforced through a carefully designed training objective. Let $p_{\text{data}}(x)$ represent the data distribution and $\mathcal{T}$ denote a time sampling strategy over $[0, 1]$. The loss function is defined as:

\begin{equation}
\begin{split}
    \mathcal{L}_{\text{consistency}} = \mathbb{E}_{x \sim p_{\text{data}}, t_1, t_2 \sim \mathcal{T}} \\ 
    \left[ \left\| \text{sg}(f_\theta(x, t_1)) - f_\theta(x, t_2) \right\|_2^2 \right]
\end{split}
\end{equation}

where $\text{sg}$ represents the stop-gradient operator. The time sampling strategy $\mathcal{T}$ is designed to focus on regions where the model's outputs are most sensitive to noise, improving learning efficiency.

\subsubsection{Efficiency Mechanism}
The remarkable efficiency of Consistency Models stems from three key innovations:

\begin{itemize}
    \item \textbf{Direct Mapping}: Instead of iterative denoising, CMs learn a direct mapping from noisy to clean images, significantly reducing computational overhead.
    
    \item \textbf{Consistency Distillation}: The model distills knowledge across different noise levels, enabling faster inference while maintaining generation quality.
    
    \item \textbf{Adaptive Sampling}: The framework allows for flexible sampling strategies, where the number of steps can be adjusted based on computational constraints without retraining.
\end{itemize}

The practical benefits are substantial, including reduced inference time compared to traditional diffusion models, lower memory requirements during both training and inference, improved stability in the generation process, and maintenance of high-quality outputs despite fewer computational steps.

Recent work has demonstrated that CMs can achieve comparable or superior performance to diffusion models while requiring significantly fewer computational resources \cite{song2023consistency}. This efficiency makes them particularly suitable for real-time applications and large-scale deployment scenarios.

\subsubsection{Extensions and Applications}
Recent developments have extended the basic CM framework in several directions:

\begin{itemize}
    \item \textbf{Conditional Generation}: Incorporating conditioning information while maintaining the consistency property
    \item \textbf{Multi-scale Generation}: Applying consistency across different spatial resolutions
    \item \textbf{Hybrid Approaches}: Combining consistency models with other generative frameworks for enhanced performance
\end{itemize}

These advances have made Consistency Models increasingly attractive for practical applications, offering a promising direction for future research in efficient image generation.

\section{Technical Advancements}

\subsection{Training Efficiency Improvements}
Advancements in image generation models are accompanied by significant improvements in training efficiency, enabling faster development and deployment of these sophisticated models while reducing computational resources and costs.

\subsubsection{Model Quantization Techniques}
Quantization has become a crucial technique for improving both training and inference efficiency of image generation models, making it possible to deploy large image generation models in resource-constrained environments with acceptable quality levels. Post-training quantization methods \cite{li2023q, shang2023post} are capable to reduce model size and memory requirements while preserving generation quality. Recent quantization-aware training approaches \cite{he2023efficientdm, chen2024efficientqat} have shown promise in maintaining high-quality image generation capabilities while operating at reduced precision. Techniques such as mixed-precision training \cite{micikevicius2017mixed} and adaptive quantization \cite{zhou2018adaptive} has also been optimized for image generation, allowing models to automatically adjust precision levels based on the importance of layers and operations. 

\subsubsection{Parameter-Efficient Fine-tuning}
PEFT, or Parameter-Efficient Fine-Tuning, is a machine learning technique designed to adapt large pre-trained models to specific tasks without needing to fine-tune all the model's parameters \cite{houlsby2019parameter}. Let $M$ be a pre-trained model with parameters $\theta \in \mathbb{R}^n$. The traditional fine-tuning approach updates all parameters, while PEFT introduces a smaller set of trainable parameters $\phi \in \mathbb{R}^m$ where $m \ll n$. The optimization objective can be expressed as:
\begin{equation}
    \min_{\phi} \mathcal{L}(f(x; \theta, \phi))
\end{equation}
where $\mathcal{L}$ is the task-specific loss function and $f$ represents the model's forward pass.

\paragraph{Low-Rank Adaptation (LoRA)}
LoRA introduces low-rank factorization into the adaptation process. For each weight matrix $W_0 \in \mathbb{R}^{d \times k}$, a low-rank update is introduced:
\begin{equation}
    W = W_0 + \Delta W = W_0 + BA
\end{equation}
where $B \in \mathbb{R}^{d \times r}$, $A \in \mathbb{R}^{r \times k}$ and $r$ is the rank (typically $r \ll \min(d,k)$) \cite{hu2021lora}. LoRA significantly reduces the number of trainable parameters while maintaining model performance. Building upon this, Quantized LoRA (QLoRA) \cite{dettmers2024qlora} further reduces memory requirements by combining low-rank adaptation with 4-bit quantization, enabling fine-tuning of large models on consumer-grade GPUs. LoRA has been particularly successful when applied to Stable Diffusion models \cite{rombach2022high}, allowing efficient creation of specialized versions that can maintain the quality of full fine-tuning while using only a fraction of the parameters \cite{li2024diffstyler}. Recent work has demonstrated that LoRA-adapted models can achieve comparable or even superior results to full fine-tuning in specific domains \cite{luo2023lcm}.

\paragraph{Other PEFT Methods}

Beyond LoRA, researchers have developed various parameter-efficient fine-tuning (PEFT) techniques specifically optimized for image generation models. Adapter-based approaches \cite{mou2024t2i, ye2023ip} insert small learnable modules while keeping the pre-trained weights frozen, offering a balance between efficiency and performance. Prefix-tuning methods \cite{hao2024optimizing} have been adapted for image generation tasks, allowing models to be specialized for different domains by learning only a small set of continuous task-specific vectors. Hybrid approaches combines multiple PEFT techniques \cite{liu2024sparsity}, demonstrating that carefully designed combinations can outperform individual methods while maintaining computational efficiency. These developments have made it possible to customize large image generation models for specific applications with minimal computational overhead.

\subsubsection{Distributed Training Strategies}
The scale of modern image generation models has necessitated sophisticated distributed training strategies. Data-parallel training has evolved with techniques like ZeRO \cite{rajbhandari2020zero} that optimize memory usage across multiple GPUs. Pipeline parallelism has been enhanced with dynamic scheduling algorithms \cite{narayanan2021efficient} that minimize device idle time and maximize throughput. A hybrid approach combines different parallelism strategies \cite{fang2024xdit} allowing efficient scaling across hundreds or thousands of GPUs while maintaining high hardware utilization. 

\section{Emerging Capabilities}

\subsection{Inpainting and Outpainting}

Inpainting reconstructs missing or corrupted regions within an image. Convolutional Neural Network (CNN)-based methods utilize convolutional layers to capture local and contextual information for filling missing regions \cite{wu2021deep}. GAN-based approaches use a generator and discriminator network to produce realistic inpainted images by learning the underlying data distribution \cite{yu2018generative}. Transformer architectures which capture long-range dependencies, have also been adapted for image inpainting \cite{lu2023grig}.

Diffusion model-based methods have been widely used in impainting tasks. Inpainting method can operate in the latent space, enabling efficient and high-quality image reconstruction \cite{corneanu2024latentpaint}. RePaint utilizes the DDPM to iteratively refine the inpainted regions, ensuring coherence with the surrounding context \cite{lugmayr2022repaint}. GradPaint introduces gradient-guided inpainting, which steers the generation process towards globally coherent images using the model's denoised image estimation \cite{grechka2024gradpaint}. T2I-Adapter integrates lightweight modules into pre-trained text-to-image models, enabling the generation of images with specific attributes by conditioning on various control signals, thereby enhancing the flexibility and precision of inpainting processes \cite{mou2024t2i}.

Outpainting, or image extrapolation, extends the original boundaries of an image by synthesizing visually coherent content that blends seamlessly with the existing regions. GAN-based methods in outpainting utilize a generator-discriminator setup to produce realistic extensions. For instance, the In\&Out method frames outpainting as a GAN inversion task, enabling diverse, semantically rich extensions by mapping input images into the GAN’s latent space \cite{cheng2022inout}. The Edge-Guided Bidirectional Outpainting technique uses transformers to guide the generation process by using boundary rearrangement and progressive learning to ensure content alignment \cite{kim2021painting}. Structural attention has also been used to guide the model’s understanding of spatial relationships, improving contextual alignment in generated images \cite{zhang2023towards}.

\subsection{Multi-view Generation}

Multi-view image generation has become essential for creating 3D-consistent visuals, enabling models to produce multiple coherent perspectives of the same scene or object. A pioneering approach in this field, MVDiffusion \cite{tang2023emergent} introduced a framework that uses cross-attention mechanisms to ensure geometric consistency across viewpoints, generating multiple views simultaneously with strong 3D awareness. Zero123 \cite{liu2023zero} enabled novel view synthesis from a single image without explicit 3D supervision by fine-tuning a pre-trained text-to-image diffusion model on multi-view datasets. This zero-shot method achieved efficient view generation through prompt engineering and careful conditioning, broadening the potential applications of multi-view synthesis. Wonder3D \cite{long2024wonder3d} advanced efficiency in multi-view generation by introducing a two-stage pipeline. It first generates a canonical view using text prompts, then a view-conditioned generation process to ensure consistency across various angles. Make-It-3D \cite{tang2023make} generates high-fidelity 3D models from a single image by leveraging 2D diffusion as a 3D-aware prior. The method constructs a neural radiance field optimized via score distillation, ensuring fidelity to the reference image, then refines textures in a point cloud stage using diffusion priors to achieve realistic, aligned geometry and appearance.

\subsection{Control and Customization}

Improvements in image generation models now enable greater control and customization, producing more accurate and reliable results. This section reviews key developments in these techniques.

\subsubsection{ControlNet}

ControlNet is a neural network architecture designed to enhance text-to-image diffusion models by incorporating additional conditioning inputs \cite{zhang2023adding}. ControlNet architecture incorporates structured conditioning elements like edge maps, depth maps, and human poses into the diffusion model pipeline to enable precise, targeted control over the image generation process. ControlNet-XS is a streamlined version of ControlNet that reinterprets the control mechanism as a feedback-control system \cite{zavadski2023controlnet}. Uni-ControlNet extends the ControlNet framework by enabling simultaneous utilization of multiple control signals, both local (e.g., edge maps, depth maps) and global (e.g., CLIP image embeddings). This approach allows for flexible and composable control within a single model, enhancing the versatility of text-to-image diffusion models \cite{zhao2024uni}.

\subsubsection{Custom Style Transfer}

Custom style transfer in image generation allows models to apply user-defined styles to images effectively. Recent advancements have introduced techniques for achieving flexible, personalized style adaptation. StyleDrop enables pre-trained text-to-image models to generate images in user-defined styles through fine-tuning based on reference images, broadening stylistic versatility \cite{sohn2023styledrop}. The Any-to-Any Style Transfer model incorporates interactive segmentation to map regions in style images to specific content areas, enabling precise regional control over style transfer through human-computer interaction \cite{liu2023any}. CLIPstyler bypasses the need for reference images by using descriptive text to apply styles via CLIP’s pre-trained embeddings, making the style transfer process more accessible \cite{kwon2022clipstyler}. StyleShot further enhances customization by training a style-aware encoder that captures various styles efficiently, supporting style transfer without the need for tuning during testing \cite{gao2024styleshot}. These methods represent a leap forward in custom style transfer, providing users with practical tools to generate distinct, high-quality results.

\subsubsection{Detail Enhancement Methods}

Recent advancements in detail enhancement methods have introduced innovative techniques to improve image quality. Jiang et al. proposed a method based on the Metropolis theorem, treating the search for optimal image features as a thermodynamic process to enhance details effectively \cite{jiang2023metropolis}. Wong presented a multi-scale image decomposition approach using a local statistical edge model, enabling progressive detail enhancement across various scales \cite{wong2021multi}. CRNet is a detail-preserving network that unifies image restoration and enhancement tasks, achieving superior performance by integrating high-frequency enhancement modules \cite{yang2024crnet}. ECAFormer is a low-light image enhancement model utilizing cross-attention mechanisms to preserve details while improving illumination \cite{ruan2024ecaformer}. These methods represent significant progress towards image detail enhancement.



\section{Performance Metrics and Evaluation}
Evaluating image generation models presents unique challenges due to the subjective nature of image quality and the multiple dimensions of performance that need to be assessed. This section explores the various metrics and methodologies employed to evaluate these models comprehensively.

\subsection{Image Quality Metrics}
Quantitative assessment of generated image quality relies on both traditional and AI-driven metrics. Fréchet Inception Distance (FID) \cite{heusel2017gans} remains a cornerstone metric, measuring the similarity between the distribution of generated ($g$) and real ($r$) images in feature space. FID is defined as:
\begin{equation}
    \text{FID}(r,g) = \|\mu_r - \mu_g\|_2^2 + \text{Tr}(\Sigma_r + \Sigma_g - 2(\Sigma_r\Sigma_g)^{1/2})
\end{equation}
$\mu_r, \mu_g \in \mathbb{R}^d$ represent the empirical means, $\Sigma_r, \Sigma_g \in \mathbb{R}^{d \times d}$ is the empirical covariance matrices of the real and generated feature distributions respectively, and $\text{Tr}$, the trace operator, computes the sum of the matrix diagonal elements. CleanFID, addresses several inconsistencies in FID implementations and providing more reliable comparisons across different studies \cite{parmar2021cleanfid}. Kernel Inception Distance (KID) \cite{binkowski2018demystifying} has been used as an alternative to FID. Instead of assuming Gaussian distributions, it measures the similarity between real and generated distributions using a polynomial kernel by computing the squared Maximum Mean Discrepancy (MMD) between the two distributions. The Learned Perceptual Image Patch Similarity (LPIPS) metric \cite{zhang2018unreasonable} tries to aligns more closely with human perception by calculating the perceptual distance between two images by comparing their feature representations from a deep neural network, typically a pre-trained network like VGG or AlexNet. Recently, more sophisticated metrics like ImageReward \cite{xu2024imagereward} have used large vision-language models to assess both image quality and text-image alignment simultaneously. \cite{tang2024clip} also proposed proposed CLIP-based metrics that better correlate with human judgments of image quality and fidelity.

\subsection{Human Evaluation Methods}
Human evaluation remains crucial for validating image generation models. \cite{petsiuk2022human} introduced a standardized framework for human evaluation campaigns, addressing common pitfalls and biases in subjective assessments, providing guidelines for large-scale human evaluations. Developing specialized platforms for human evaluation \cite{turchi2023human, lin2025evaluating, sun_qin_peng_2024} enables more efficient and reliable collection of human judgments. These platforms incorporate mechanisms for quality control, annotator training, and bias detection.

\subsection{Prompt Alignment Metrics}
Evaluating text-image alignment has been a critical aspect of model assessment. CLIP-based metrics \cite{hessel2021clipscore} have become standard for measuring prompt-image correspondence, offering automated assessment of how well generated images match their text descriptions. Semantic fidelity metrics \cite{zhang2024enhancing} focus on evaluating how well specific concepts and attributes from the prompt are preserved in the generated image. These metrics decompose prompts into semantic units and assess their presence in the generated output. Besides, \cite{baryshnikov2023hypernymy} introduced hierarchical prompt evaluation methods that consider both high-level concepts and fine-grained details.

\subsection{Computational Efficiency Metrics}
As image generation models grow in complexity, evaluating computational efficiency becomes increasingly important. Standard metrics include throughput (images per second), memory usage, and FLOPs. Memory-aware metrics \cite{ryu2025memory} provide detailed analysis of memory utilization patterns, helping identify bottlenecks and optimization opportunities. These metrics consider peak memory usage, memory fragmentation, and cache efficiency. MLPerf \cite{reddi2020mlperf} introduced standardized benchmarks for comparing different architectures' efficiency across various hardware configurations. The growing focus on environmental impact has led to the development of energy consumption metrics \cite{patterson2021carbon}, measuring the carbon footprint of model training and inference.

\section{Future Directions}
\subsection{Current Limitations}

Despite advancements in image generation, key technical challenges remain. Generating high-quality images with prompts containing multiple components often proves difficult, as models struggle to incorporate all visual elements, leading to reduced quality and context accuracy with increased complexity \cite{foong2023challenges}. Integrating human preferences into generation is also an active area of research. Tools like PrefPaint \cite{liu2024prefpaint} have emerged to align models with human aesthetic standards, yet fully capturing and embedding these preferences remains complex. These barriers underscore the ongoing need for research to enhance both the performance and usability of image generation models.

\subsubsection{Resource Constraints}

Deep generative image models require significant computational resources during both training and inference, posing a major barrier to accessibility and scalability. Training state-of-the-art models involves processing massive datasets across thousands of GPUs or TPUs over extended periods, which is often prohibitively expensive for smaller research groups and organizations. Additionally, inference in these models is computationally intensive, particularly when generating high-resolution or highly detailed images. These resource demands exacerbate inequities in access to cutting-edge technologies and hinder broader adoption in resource-constrained environments. Future research must prioritize model efficiency, including techniques like quantization, pruning, and knowledge distillation, to reduce the computational footprint without compromising quality.

\subsubsection{Quality Challenges}

Despite advancements, deep generative image models still face substantial challenges in generating consistent and high-quality outputs. When prompts become more complex, models often fail to accurately capture all specified attributes, resulting in incomplete or misaligned visual representations. Issues such as artifacts, unrealistic textures, and poor compositional coherence are prevalent in edge cases or underrepresented data distributions. Aligning generated images with nuanced human aesthetic standards remains difficult. Addressing these challenges will require additional work in architecture design, training methodologies, and evaluation metrics. Emphasis on integrating diverse, high-quality datasets and utilizing multimodal feedback systems could significantly enhance the quality and fidelity of generated images.

\subsection{Promising Research Areas}

Aesthetic quality control in image generation has seen promising developments aimed at enhancing visual appeal. Playground v2.5 \cite{li2024playground} refines color, contrast, and human-centric details in text-to-image models, achieving top-tier aesthetic results. AIGCIQA2023 \cite{wang2023aigciqa2023} introduced a large-scale database that evaluates AI-generated images on quality, authenticity, and correspondence to human preferences, providing a structured benchmark for aesthetic assessment. Emu \cite{dai2023emu}, a model fine-tuned with high-quality images, guides pre-trained models to produce visually appealing images, demonstrating superior results in aesthetic quality. These advancements collectively mark aesthetic quality control as a rapidly evolving research area with significant impact on the field.

Prompt engineering has been important in enhancing generative image models. PromptMagician \cite{feng2023promptmagician} provides an interactive system for refining input prompts, analyzing generated images to suggest modifications that better align with user intent. BeautifulPrompt \cite{cao2023beautifulprompt} generates high-quality prompts from basic descriptions, using reinforcement learning with visual AI feedback to optimize both prompt quality and the aesthetic appeal of generated images. NeuroPrompts \cite{rosenman2023neuroprompts} adapts user prompts through constrained text decoding with a pre-trained language model, producing images that more closely meet user expectations. These innovations demonstrate the potential of prompt engineering to improve generative models, emphasizing the role of adaptive systems and user-friendly interfaces in bridging the gap between user intent and model output.

Ensuring the safety of generative image models is critical \cite{peng2024jailbreaking, peng2024securing}. Studies reveal potential misuse risks, including harmful content generation. SneakyPrompt \cite{yang2024sneakyprompt} exposed vulnerabilities in model safety by "jailbreaking" text-to-image models, while GuardT2I \cite{yang2024guardt2i} was introduced to counter such threats, using large language models to detect and neutralize adversarial prompts. Additionally, UnsafeBench \cite{qu2024unsafebench} benchmarks image safety classifiers on both real-world and AI-generated content, providing a rigorous assessment of existing safety protocols. These advancements highlight the ongoing need for robust safety measures to prevent misuse and promote responsible deployment of generative image technology.

Research in generative image models has also focused on developing efficient architectures that balance computational demands with high-quality output. TurboViT \cite{wong2023turbovit} exemplifies this by using generative architecture search to create a vision transformer that is both compact and high-performing. ControlNet-XS \cite{zavadski2023controlnet} introduces an optimized architecture for controlling text-to-image diffusion models, achieving efficiency without sacrificing image quality. For ultra-high-resolution synthesis, the Pyramid Diffusion Model (PDM) \cite{yang2024ultra} employs a pyramid latent representation to generate scalable images up to 2K resolution. These innovations highlight the critical role of architectural advancements in creating resource-efficient models that deliver high-fidelity results.

\subsection{Conclusion}
This review has traced the evolution of deep generative image models, from early GANs to modern diffusion and transformer-based architectures. Innovations in latent representations, consistency models, and hybrid approaches have dramatically improved image quality while reducing computational demands. PEFT and advanced quantization techniques have made these models more accessible and deployable.

Although the field has advanced considerably in efficiency and quality, fundamental challenges in scalability and optimization persist. The path forward lies in developing efficient architectures and human-aligned evaluation methods, with increasing focus on responsible deployment that considers both technical and societal dimensions.

\bibliographystyle{ieeetr}  
\bibliography{references} 


\end{document}